\documentclass[runningheads]{llncs}
\usepackage[T1]{fontenc}
\usepackage{amsmath, amssymb}
\usepackage{algorithm}
\usepackage{algpseudocode}
\usepackage{graphicx}
\usepackage{booktabs}
\usepackage[misc]{ifsym}
\usepackage{tikz}
\usetikzlibrary{shapes, arrows, positioning, fit, calc}

\begin{document}

\title{Decomposing Observational Multiplicity in Decision Trees: Leaf and Structural Regret}

\titlerunning{Decomposing Observational Multiplicity in Decision Trees}

\author{Mustafa Cavus\orcidID{0000-0002-6172-5449}}

\authorrunning{Cavus}

\institute{Eskisehir Technical University, Department of Statistics, Turkiye 26555 \email{mustafacavus@eskisehir.edu.tr}}

\maketitle 
\begin{abstract}
Many machine learning tasks admit multiple models that perform almost equally well, a phenomenon known as predictive multiplicity. A fundamental source of this multiplicity is \textit{observational multiplicity}, which arises from the stochastic nature of label collection: observed training labels represent only a single realization of the underlying ground-truth probabilities. While theoretical frameworks for observational multiplicity have been established for logistic regression, their implications for non-smooth, partition-based models like decision trees remain underexplored. In this paper, we introduce two complementary notions of observational multiplicity for decision tree classifiers: \textit{leaf regret} and \textit{structural regret}. Leaf regret quantifies the intrinsic variability of predictions within a fixed leaf due to finite-sample noise, while structural regret captures variability induced by the instability of the learned tree structure itself. We provide a formal decomposition of observational multiplicity into these two components and establish statistical guarantees. Our experimental evaluation across diverse credit risk scoring datasets confirms the near-perfect alignment between our theoretical decomposition and the empirically observed variance. Notably, we find that structural regret is the primary driver of observational multiplicity, accounting for over 15 times the variability of leaf regret in some datasets. Furthermore, we demonstrate that utilizing these regret measures as an abstention mechanism in selective prediction can effectively identify arbitrary regions and improve model safety, elevating recall from 92\% to 100\% on the most stable sub-populations. These results establish a rigorous framework for quantifying observational multiplicity, aligning with recent advances in algorithmic safety and interpretability.
\keywords{Observational multiplicity \and Predictive multiplicity \and Decision trees \and Regret decomposition \and Selective prediction.}
\end{abstract}
\section{Introduction}

The deployment of machine learning models in high-stakes domains---such as healthcare, credit scoring, and legal risk assessment---demands not only high prediction performance but also reliable predictions at the individual level. Nevertheless, a growing body of work has demonstrated that many learning problems admit \emph{predictive multiplicity}: the existence of multiple models that achieve near-identical accuracy while assigning conflicting predictions to the same individual \cite{breiman_2001,marx_et_al_2020}. This phenomenon implies that an individual prediction may depend on the arbitrary selection of one model from this set of equally valid alternatives, giving rise to what has been termed \emph{predictive arbitrariness} \cite{black_et_al_2022}. Such arbitrariness is particularly problematic in high-stakes decision-making, where model outputs must be justifiable not only in aggregate but also at the level of individual decisions.

Early explanations of predictive multiplicity primarily emphasized sources of uncertainty arising from model underspecification, including architectural choices and optimization pipelines \cite{damour_et_al_2022}. While this line of work has shown that modern machine learning workflows can produce many empirically indistinguishable models, it does not fully account for uncertainty originating from the data-generating process itself. More recently, \emph{observational multiplicity} has been identified as a fundamental and distinct source of predictive multiplicity \cite{george_et_al_2025}. This phenomenon captures the idea that observed binary labels are stochastic realizations of latent ground-truth probabilities. Consequently, different but equally plausible draws of labels from the same underlying distribution may lead to different trained models. This irreducible label-induced variance is closely related to the notion of \emph{dataset multiplicity}, which explores how model outcomes vary across all \textit{nearby} versions of the data under uncertainties like label errors or bias.

Despite recent theoretical advances, the implications for non-smooth models like decision trees remain underexplored. Decision trees are notoriously sensitive to data perturbations, behaving similarly to nearest-neighbor classifiers in terms of stability. Classic stability theory highlights this: for example, a tree with $v$ leaves has a defined hypothesis stability $\beta_h$, implying that very small or shallow trees exhibit poor stability compared to deeper structures. This dual nature suggests two distinct sources of predictive multiplicity: label noise within a fixed tree leaf, representing irreducible \emph{aleatoric uncertainty}, and variation across different tree structures, representing \emph{epistemic uncertainty} arising from model change.

In this work, we bridge this gap by formalizing a decomposition of observational multiplicity into leaf and structural regret, providing a practical framework to distinguish between local prediction noise and global model instability---a distinction crucial for deploying safe, tree-based systems in high-stakes environments. Our primary contribution is the introduction of two complementary notions of regret:
\begin{itemize}
    \item \textbf{Leaf Regret:} Quantifies the intrinsic variability of predictions within a fixed leaf, conditional on a given tree structure, capturing uncertainty due to stochastic label realizations.
    \item \textbf{Structural Regret:} Captures the additional variability induced by randomness in the learned tree structure itself, reflecting instability across equally plausible training label realizations.
\end{itemize}

Through this decomposition, we provide a more granular understanding of how predictive multiplicity propagates in non-smooth models. We establish theoretical guarantees and concentration inequalities for estimating leaf regret, and propose Monte Carlo procedures to approximate structural regret. By leveraging these regret measures, we demonstrate how decision systems can proactively identify individuals for whom the model is ``guessing'' due to label instability, thereby enabling a more rigorous approach to algorithmic safety. Our framework contributes to the growing literature on algorithmic safety and interpretability by explicitly distinguishing between \emph{noise within a structure} and \emph{instability of the structure} in tree-based learning.

The remainder of this paper is organized as follows. Section~2 reviews related work. Section~3 presents our methodology and theoretical results. Section~4 reports experimental evaluations. Section~5 discusses implications and limitations, and Section~6 concludes.


\section{Related Work}

The reliability of machine learning models has been increasingly scrutinized through the lens of \textit{predictive multiplicity}, a phenomenon where multiple models with near-identical aggregate performance assign conflicting predictions to the same individual \cite{marx_et_al_2020}. This challenge, often referred to as the \textit{Rashomon effect} \cite{breiman_2001}, poses significant implications for interpretability, fairness, and safety in high-stakes decision-making \cite{damour_et_al_2022,black_et_al_2022}. Modern machine learning pipelines typically fit models through empirical risk minimization, which may return multiple competing models that differ in their individual-level predictions despite achieving the same empirical risk. While traditional approaches viewed this as an optimization or underspecification problem, recent studies have highlighted the ethical concerns of predictive arbitrariness, where an individual's treatment depends on the arbitrary choice of one optimal model over another from the Rashomon set.

A critical and unavoidable source of this arbitrariness is \textit{observational multiplicity}, which describes the uncertainty arising from the stochastic nature of data collection \cite{george_et_al_2025}. Unlike the optimization-driven multiplicity discussed by D'Amour et al. \cite{damour_et_al_2022}, which stems from model underspecification, observational multiplicity considers uncertainty as a byproduct of the data-generating process itself. This distinction is vital: while the Rashomon effect explores the space of equally performing models, observational multiplicity quantifies the unavoidable variability induced by the stochastic realization of training labels. This perspective notes that training labels are often single realizations of underlying ground-truth probabilities; consequently, different but equally plausible draws of labels from the same distribution would lead to different trained models. To quantify this irreducible risk, the measure of \textit{regret} \cite{george_et_al_2025} is proposed to capture unavoidable variability in individual predictions due to randomness in observed labels. This concept is closely related to the \textit{dataset multiplicity} problem \cite{meyer_2023}, which explores how unreliable data and counterfactual label draws impact predictions. In parallel, probabilistic formulations of multiplicity show that disagreement can persist even when models are calibrated and optimized on proper losses \cite{watsondaniels_2022}, while Rashomon-capacity analyses provide quantitative characterizations of the breadth of near-optimal model sets \cite{hsu_calmon_2022}. At a broader decision level, multi-target multiplicity further shows that conflicts can propagate across multiple outcomes in high-stakes settings \cite{watsondaniels_2023}. A key limitation in existing regret-based frameworks is their primary focus on smooth, differentiable classifiers. While George et al. \cite{george_et_al_2025} introduce measures of regret to capture irreducible risk, their theoretical assumptions often do not account for the discrete instability inherent in partition-based models. Our work bridges this gap by addressing how observational multiplicity manifests through discontinuous decision boundaries. Beyond the inherent stochasticity of labels, the broader data-centric pipeline also shapes these effects; for instance, the common preprocessing techniques, such as class balancing and filtering, can inadvertently inflate predictive multiplicity, further complicating the Rashomon effect \cite{cavus_biecek_2025}. Both perspectives agree that certain examples, such as boundary cases or outliers, are more susceptible to conflicting predictions across plausible versions of the data. Our work lies at this intersection, studying how such label-driven and preprocessing-augmented uncertainty propagates specifically through partition-based models.

Decision trees are notoriously sensitive to small perturbations in training data, a property characterized as \textit{algorithmic instability} \cite{bousquet_and_elisseeff_2002}. Recent derivations show that a tree's stability is tightly linked to its complexity, where increasing the number of leaves or depth can rapidly alter stability properties \cite{arsov_2019}. While stability analysis has been used to quantify how changes in datasets propagate to the learned model, it often fails to distinguish between different sources of variance in non-smooth classifiers. Although ensemble methods like bagging and boosting are commonly used to mitigate this sensitivity, they do not isolate the specific source of variance within the tree structure itself. Traditional stability analysis, such as the framework provided in \cite{bousquet_and_elisseeff_2002}, typically assesses how changes in the dataset propagate to the global model output but fails to decompose the specific origins of this variance. By formalizing the decomposition into leaf and structural components, we provide a more granular lens that distinguishes between localized aleatoric noise and global epistemic instability. Our work extends this literature by providing a formal decomposition of this variability into \textit{leaf regret} (within-structure noise) and \textit{structural regret} (between-structure instability), bridging the gap between high-level observational multiplicity theory and classical tree-based induction.

Furthermore, the impact of noisy labels remains a pervasive issue in supervised learning. While loss functions like mean absolute error are known to be noise-robust, standard cross-entropy is not \cite{ghosh_2017}. Theoretical evidence suggests that standard CART splits remain robust under symmetric label noise in large datasets, yet modern deep-learning noise-correction methods offer limited benefit for decision tree models \cite{sztukiewicz_2024}. This resilience to standard noise-correction highlights a fundamental characteristic of partition-based induction: tree structures respond to label fluctuations not just through bias, but through significant structural re-orientation. In complementary lines of safety-oriented work, selective classification and reject-option frameworks demonstrate how abstention can improve reliability under uncertainty \cite{chow_1970,geifman_2017}. Consequently, identifying the arbitrariness of a decision requires a diagnostic tool that can attribute uncertainty to either local leaf-level variance or global structural shifts. This indicates a pressing need for specialized tree-robust techniques that account for the unique way partition-based models react to label fluctuations. Our framework contributes to this need by providing a formal method to attribute variance to its specific source---either local leaf variability or global structural instability---in noisy classification settings.
\section{Methodology}

This section formalizes two complementary notions of \textit{observational multiplicity} for decision tree classifiers: \textit{leaf regret} and \textit{structural regret}. We decompose total predictive uncertainty into within-structure variability---captured by leaf regret under a fixed partition---and between-structure instability, which arises when the tree structure itself varies due to stochastic label realizations. Our analysis proceeds in two stages: first, we establish precise statistical guarantees and concentration inequalities for leaf regret by treating the tree structure as fixed; second, we relax this constraint to investigate structural regret as a manifestation of algorithmic instability under observational multiplicity.

\subsection{Setup and notation}

Let $\mathcal{D} = \{(X_i, Y_i)\}_{i=1}^n$ be a training dataset where $X_i \in \mathcal{X}$ and $Y_i \in \{0,1\}$. In probabilistic classification, we assume the relationship between features and labels is inherently stochastic, such that $Y_i$ represents a single realization of an underlying ground-truth probability $p_i^* := \mathbb{P}(Y_i = 1 \mid X_i)$. 

This setting gives rise to \textit{observational multiplicity}: the phenomenon where different, equally plausible draws of labels from the same underlying distribution $\mathbb{P}(Y \mid X)$ would result in different trained models that may assign conflicting predictions to the same individual. We formalize the uncertainty arising from this effect through the lens of \textit{regret}, which captures the unavoidable variability in predictions due to the randomness in the observed labels.

Specific to tree-based models, let $L$ denote a fixed leaf of a decision tree trained on $\mathcal{D}$. Conditioned on the event that an observation falls into leaf $L$, we assume $Y_i \mid X_i \in L \sim \mathrm{Bernoulli}(p_L^\ast)$, where $n_L$ is the number of observations in leaf $L$. This assumption aligns with the standard interpretation of classification trees, where each leaf induces a constant class probability estimate. By conditioning on a fixed leaf, we ensure that all observations are evaluated within the same partition cell induced by the tree, such that the remaining randomness arises solely from the response variable---a special case of observational multiplicity within a fixed structure.

\subsection{Leaf regret}

We define \emph{leaf regret} as the conditional variance of the leaf-level probability estimator, $R_L^{\mathrm{leaf}} := \mathrm{Var}(\hat p_L \mid L)$, where the leaf-level probability estimator is $\hat p_L = 1 / n_L \sum_{i \in L} Y_i$. It measures the intrinsic instability of predictions within leaf $L$ due to finite-sample variability.

\begin{lemma}[Well-definedness of leaf regret]
\label{lemma1}
For any leaf $L$ with $n_L \ge 1$, the quantity $R_L^{\mathrm{leaf}} := \mathrm{Var}(\hat p_L \mid L)$ is finite and admits the closed-form expression $R_L^{\mathrm{leaf}} = \frac{p_L^\ast (1 - p_L^\ast)}{n_L}$.
\end{lemma}

\noindent Lemma~\ref{lemma1} establishes that leaf regret is a well-defined statistical quantity with an explicit expression, rather than an ad hoc heuristic.

\begin{lemma}[Uniform upper bound]
\label{lemma2}
For any leaf $L$, $R_L^{\mathrm{leaf}} \le \frac{1}{4n_L}$.
\end{lemma}

\noindent The uniform bound in Lemma~\ref{lemma2} shows that leaf regret is maximized when class uncertainty is highest and decreases at a rate inversely proportional to the leaf size. This highlights the central role of $n_L$ in controlling predictive instability at the leaf level.

\subsection{Estimation of leaf regret}

In practice, the true leaf probability $p_L^\ast$ is unknown and must be estimated from data. We therefore consider the empirical plug-in estimator
\begin{equation}
    \widehat R_L^{\mathrm{leaf}} :=
    \frac{\hat p_L (1 - \hat p_L)}{n_L}.    
\end{equation}

\begin{lemma}[Consistency of the plug-in estimator]
\label{lemma3}
The empirical leaf regret estimator satisfies $\widehat R_L^{\mathrm{leaf}} \xrightarrow[]{\mathbb{P}} R_L^{\mathrm{leaf}} \quad \text{as } n_L \to \infty$.
\end{lemma}

\noindent Lemma~\ref{lemma3} justifies the use of the plug-in estimator by showing that it consistently estimates the true leaf regret as the number of observations within a leaf increases.

\begin{lemma}[Deviation inequality]
\label{lemma4}
For any $\varepsilon > 0$,
\begin{equation}
    \mathbb{P}\left(\left| \hat p_L - p_L^\ast \right| > \varepsilon \right) \le 2 \exp(-2 n_L \varepsilon^2).
\end{equation}

Consequently, there exists a constant $C > 0$ such that
\begin{equation}
    \mathbb{P}\left(\left| \widehat R_L^{\mathrm{leaf}} - R_L^{\mathrm{leaf}} \right| > \delta \right) \le 2 \exp(- C n_L).
\end{equation}
\end{lemma}

\noindent Lemma~\ref{lemma4} provides a concentration guarantee, showing that the estimated leaf regret is sharply concentrated around its population counterpart. As a result, large deviations of the estimated regret become exponentially unlikely as the leaf size grows.

\subsection{Asymptotic behavior}

\begin{theorem}[Asymptotic vanishing of leaf regret]
\label{theorem1}
Assume that for a sequence of trees indexed by sample size $n$, the leaf size satisfies $n_L \to \infty$. Then, $R_L^{\mathrm{leaf}} \to 0 \quad \text{and} \quad \widehat R_L^{\mathrm{leaf}} \to 0 \quad \text{in probability}$.
\end{theorem}

\noindent Theorem~\ref{theorem1} shows that leaf regret is a finite-sample phenomenon that vanishes asymptotically as more data accumulate within a leaf. Importantly, this result does not imply improved predictive accuracy, but rather an increase in the stability of the leaf-level probability estimate.

\begin{corollary}[Expected leaf regret bound]
\label{corollary1}
If leaf $L$ is random, then
\[
\mathbb{E}[R_L^{\mathrm{leaf}}] \le \frac{1}{4}\,\mathbb{E}\!\left[\frac{1}{n_L}\right].
\]
\end{corollary}

\noindent Corollary~\ref{corollary1} links leaf regret to the distribution of leaf sizes induced by the tree structure. This connection provides a statistical interpretation of regularization mechanisms such as minimum leaf size constraints and pruning.

\subsection{Monte Carlo approximation of leaf regret}

The results above characterize leaf regret as a population-level quantity and justify the use of the plug-in estimator $\widehat R_L^{\mathrm{leaf}} = \hat p_L (1-\hat p_L)/n_L$. In practice, however, the variability of $\hat p_L$ may also be approximated numerically via Monte Carlo resampling. This approach is particularly useful when analytic expressions are unavailable or when assessing finite-sample behavior. Conditioned on the empirical leaf probability $\hat p_L$, consider the Monte Carlo procedure that generates $B$ independent samples $Y_{i}^{(b)} \sim \mathrm{Bernoulli}(\hat p_L)$, where $i = 1,\dots,n_L$, and $b = 1,\dots,B$ and computes $\hat p_L^{(b)} = \frac{1}{n_L} \sum_{i=1}^{n_L} Y_i^{(b)}$. The Monte Carlo estimator of leaf regret is then defined as

\begin{equation}
    \widehat R_{L,B}^{\mathrm{MC}} = \frac{1}{B-1} \sum_{b=1}^B \left( \hat p_L^{(b)} - \bar p_L^{(B)} \right)^2,
\end{equation}

\noindent where $\bar p_L^{(B)} = \frac{1}{B} \sum_{b=1}^B \hat p_L^{(b)}$. This estimator approximates the variance of resampled leaf-probability estimates under the plug-in Bernoulli model parameterized by $\hat p_L$, and therefore targets the plug-in estimator $\widehat R_L^{\mathrm{leaf}}$ rather than the unknown population quantity $R_L^{\mathrm{leaf}}$ directly.

\begin{lemma}[Consistency of the Monte Carlo estimator]
\label{lemma5}
Conditioned on $\hat p_L$, the Monte Carlo estimator satisfies $\widehat R_{L,B}^{\mathrm{MC}} \xrightarrow[]{\mathbb{P}} \widehat R_L^{\mathrm{leaf}} \quad \text{as } B \to \infty$.
\end{lemma}

\noindent Lemma~\ref{lemma5} follows from the law of large numbers applied to the empirical variance of $\hat p_L^{(b)}$ and ensures that the Monte Carlo procedure introduces no additional asymptotic bias beyond that of the plug-in estimator.

\begin{theorem}[Two-stage convergence of Monte Carlo leaf regret]
\label{theorem2}
If $n_L \to \infty$ and $B \to \infty$, then $\widehat R_{L,B}^{\mathrm{MC}}
\xrightarrow[]{\mathbb{P}} R_L^{\mathrm{leaf}}$.
\end{theorem}

\noindent Theorem~\ref{theorem2} establishes that the Monte Carlo estimator converges to the true leaf regret through a two-stage mechanism: statistical consistency of the plug-in estimator as the leaf size grows, and numerical consistency of the Monte Carlo approximation as the number of repetitions increases. Importantly, this result reflects convergence in estimation precision rather than improvements in predictive accuracy.

Algorithm~\ref{alg:mc_leaf_regret} summarizes the Monte Carlo procedure used
to approximate leaf regret in finite samples.

\begin{algorithm}[H]
\caption{Monte Carlo estimation of leaf regret}
\label{alg:mc_leaf_regret}
\begin{algorithmic}[1]
\Require Leaf sample $\{Y_i\}_{i=1}^{n_L}$, number of replications $B$
\Ensure Monte Carlo estimate $\widehat R_{L,B}^{\mathrm{MC}}$
\State Compute empirical leaf probability $\hat p_L = 1 / n_L \sum_{i=1}^{n_L} Y_i$
\For{$b = 1,\dots,B$}
    \State Generate $Y_i^{(b)} \sim \mathrm{Bernoulli}(\hat p_L)$ for $i=1,\dots,n_L$
    \State Compute $\hat p_L^{(b)} = \frac{1}{n_L} \sum_{i=1}^{n_L} Y_i^{(b)}$
\EndFor
\State Compute $\widehat R_{L,B}^{\mathrm{MC}} = 1/(B-1) \sum_{b=1}^B \left( \hat p_L^{(b)} - \bar p_L^{(B)} \right)^2$
\State \Return $\widehat R_{L,B}^{\mathrm{MC}}$
\end{algorithmic}
\end{algorithm}

\subsection{Structural regret}

Leaf regret quantifies the intrinsic variability of predictions within a fixed leaf, conditional on a given tree structure. In contrast, decision tree models also exhibit variability due to randomness in the induced tree structure itself, arising from sampling variability and instability of split selection. We refer to this second source of uncertainty as \emph{structural regret}.

Let $\mathcal{T}$ denote a random decision tree induced by a learning algorithm trained on a random sample from the data-generating process. For a fixed input $x$, let $L(x;\mathcal{T})$ denote the terminal node of tree $\mathcal{T}$ to which $x$ is assigned, and define the corresponding prediction $\hat p(x;\mathcal{T}) := \hat p_{L(x;\mathcal{T})}$.

\begin{definition}[Structural regret]
The structural regret at input $x$ is defined as $R^{\mathrm{struct}}(x)
:= \mathrm{Var}_{\mathcal{T}} \bigl( \hat p(x;\mathcal{T}) \bigr)$, where the variance is taken with respect to the randomness in the tree construction.
\end{definition}

Unlike leaf regret, structural regret depends on the stability properties of the tree induction algorithm and the data-generating process. As a result, its distributional form is generally intractable and algorithm-dependent. Consequently, we do not pursue plug-in or concentration-based guarantees analogous to those derived for leaf regret. Instead, structural regret is analyzed through stability arguments that characterize its asymptotic behavior under increasing sample size.

\begin{lemma}[Decomposition of predictive variability]
\label{lemma6}
For a fixed input $x$, the total predictive variability admits the decomposition
\begin{equation}
    \mathrm{Var}(\hat p(x)) = \mathbb{E}_{\mathcal{T}} \bigl[ R^{\mathrm{leaf}}_{L(x;\mathcal{T})} \bigr] + R^{\mathrm{struct}}(x),
\end{equation}

\noindent where the first term corresponds to expected leaf regret and the second term to structural regret.
\end{lemma}

\noindent Lemma~\ref{lemma6} shows that leaf regret and structural regret quantify complementary sources of predictive instability: the former arises from finite-sample variability within leaves, while the latter reflects instability of the tree partition itself.

\begin{theorem}[Vanishing structural regret under stability]
\label{theorem3}
If the tree learning algorithm is stable in the sense that $\hat p(x;\mathcal{T}) \xrightarrow[]{\mathbb{P}} p^\ast(x) \quad \text{as } n \to \infty$, then $R^{\mathrm{struct}}(x) \to 0$.
\end{theorem}

\noindent Theorem~\ref{theorem3} formalizes the intuition that structural regret vanishes when the tree structure becomes stable under increasing sample size. This result highlights the role of regularization, pruning, and ensemble methods in controlling structural instability.

Algorithm~\ref{alg:mc_structural_regret} provides a practical Monte Carlo
procedure for approximating structural regret via resampling-induced tree
instability.

\begin{algorithm}[H]
\caption{Monte Carlo estimation of structural regret}
\label{alg:mc_structural_regret}
\begin{algorithmic}[1]
\Require Training dataset $\mathcal{D}$, fixed input $x$, number of trees $B$
\Ensure Monte Carlo estimate $\widehat R^{\mathrm{struct}}_B(x)$
\For{$b = 1,\dots,B$}
    \State Draw bootstrap sample $\mathcal{D}^{(b)}$ from $\mathcal{D}$
    \State Train tree $\mathcal{T}^{(b)}$ on $\mathcal{D}^{(b)}$
    \State Compute prediction $\hat p^{(b)}(x) = \hat p(x;\mathcal{T}^{(b)})$
\EndFor
\State Compute $\widehat R^{\mathrm{struct}}_B(x) = 1/(B-1) \sum_{b=1}^B \left( \hat p^{(b)}(x) - \bar p^{(B)}(x) \right)^2$

\State \Return $\widehat R^{\mathrm{struct}}_B(x)$
\end{algorithmic}
\end{algorithm}

Together, these results establish leaf regret as a statistically well-behaved measure of local predictive instability. The guarantees derived here concern the reliability of regret estimation rather than model correctness, aligning with recent work on observational multiplicity and interpretability in tree-based models.

\section{Experiments} 

In this section, we evaluate the proposed regret decomposition framework across multiple real-world datasets. Our primary objective is to verify the mathematical identity established in Lemma 6 and quantify the relative impact of structural instability on individual-level predictive arbitrariness. The experiments are organized in two blocks: we first validate the decomposition on three datasets (\texttt{bank\_marketing}, \texttt{hmeq}, \texttt{taiwan\_credit}), then evaluate selective prediction on an expanded six-dataset benchmark that additionally includes \texttt{german\_credit}, \texttt{loan\_data}, and \texttt{poland\_credit}.
\subsection{Numerical Validation of Lemma 6}

To demonstrate the validity of our decomposition, we conducted a semi-synthetic validation study following the protocol described in \cite{george_et_al_2025}. We utilized three diverse datasets: \texttt{bank\_marketing} \cite{bank_marketing_dataset}, \texttt{hmeq} \cite{hmeq_dataset}, and \texttt{taiwan\_credit} \cite{taiwan_dataset}. For each dataset, we estimated the ground-truth probabilities using a logistic regression oracle to ensure a realistic relationship between features and labels.

We performed a nested Monte Carlo simulation by generating 200 realizations of training labels. For each realization, a decision tree was induced to estimate both structural and leaf components of regret. We compared the \textit{estimated regret} (the sum of expected leaf regret and structural regret) against the \textit{true regret} obtained through direct variance estimation over the simulations.

\begin{figure}[ht]
    \centering
    \includegraphics[width=\textwidth]{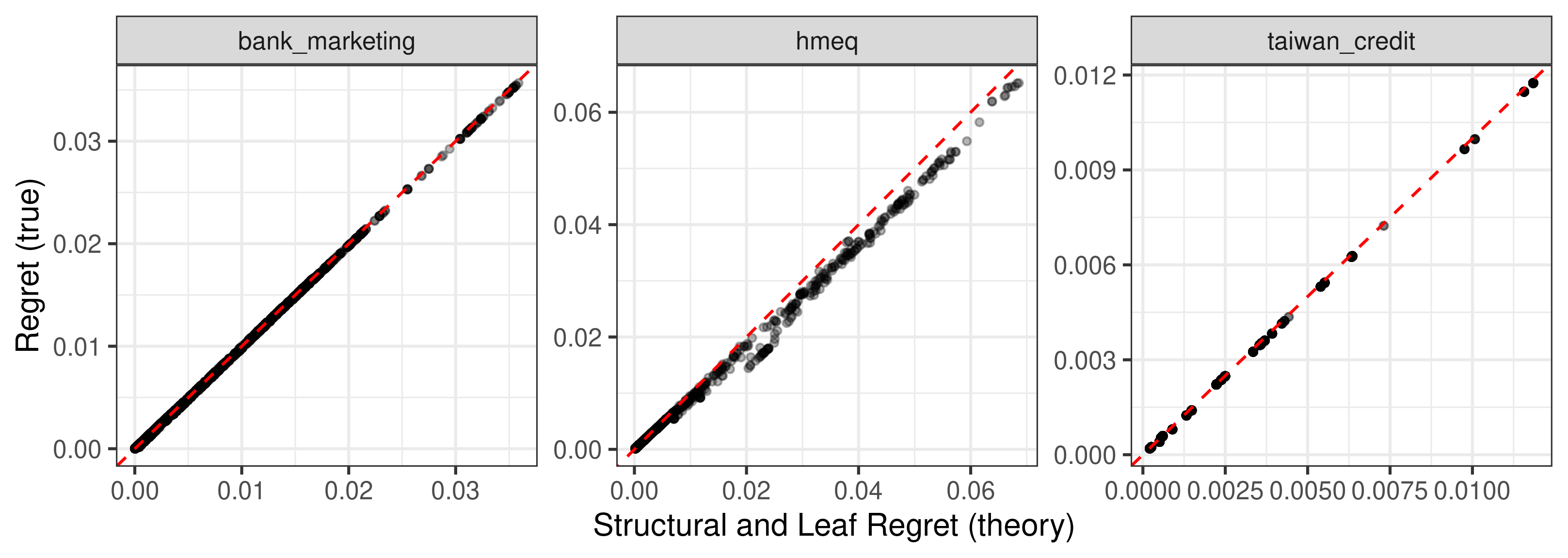}
    \caption{Theorized versus actual regret in decision trees across three datasets. The x-axis represents the sum of expected leaf regret and structural regret, while the y-axis shows the simulated true variance.}
    \label{fig:lemma6_validation}
\end{figure}

We observe a near-perfect correspondence between the estimated regret components and the actual predictive variance across all three datasets in Figure~\ref{fig:lemma6_validation}. This tight alignment along the $y=x$ line confirms that our decomposition precisely captures the constituent sources of predictive instability. This result demonstrates that even for non-smooth, partition-based classifiers, observational multiplicity can be exactly decomposed into local leaf noise and global structural instability.

\subsection{Empirical Validation of Lemma 2 and Theorem 1}

\begin{figure}[ht]
    \centering
    \includegraphics[width=\textwidth]{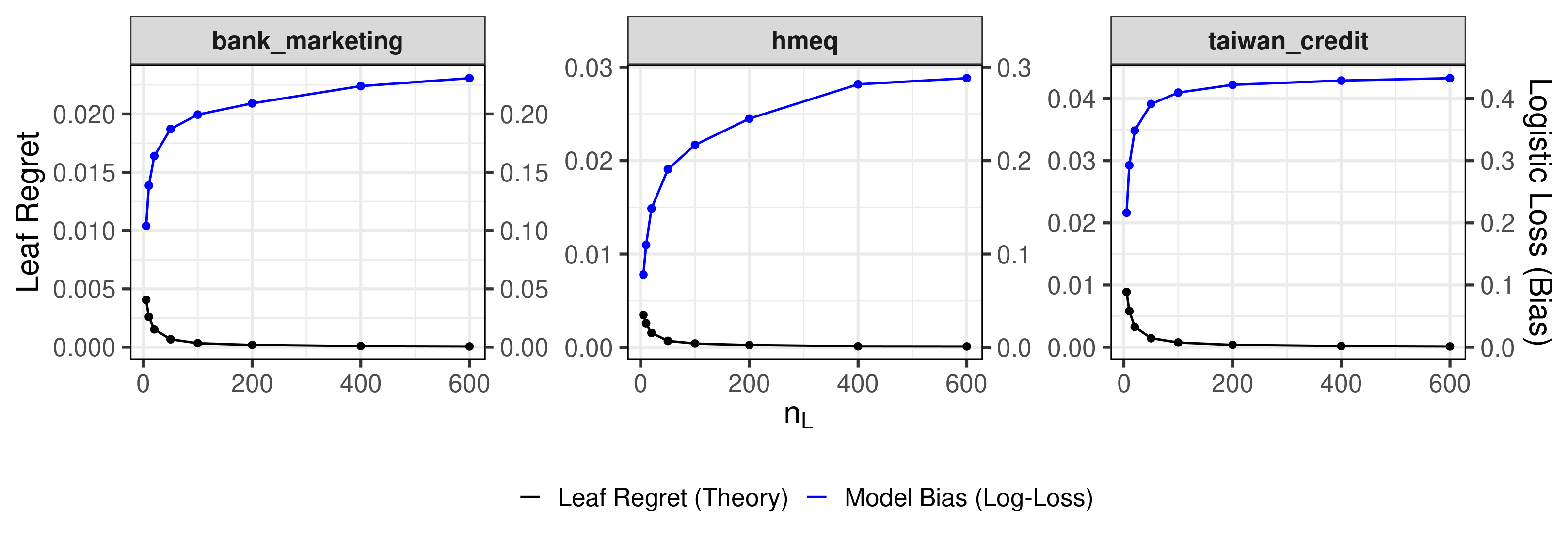}
    \caption{The impact of minimum leaf size ($n_L$) on predictive stability and model performance. The dual-axis plot illustrates the trade-off between Leaf Regret and Logistic Loss. These empirical results confirm that increasing partition size effectively mitigates observational multiplicity, as predicted in Lemma 2, while the increase in Logistic Loss reflects an empirical underfitting trade-off.}
    \label{fig:leaf_size_tradeoff}
\end{figure}

We investigated the impact of the minimum leaf size parameter ($n_L$) on the two primary components of predictive stability: local uncertainty and global performance. As illustrated in Figure~\ref {fig:leaf_size_tradeoff}, our results demonstrate a clear trade-off that aligns with our theoretical propositions.

The sharp decline in Leaf Regret as $n_L$ increases confirms \emph{Lemma 2}, showing that larger partitions effectively mitigate the impact of label-redrawing noise within local structures. Conversely, the steady rise in Logistic Loss (red dashed line) indicates an empirical underfitting effect of over-smoothing. This trend complements \emph{Theorem 1}, which concerns asymptotic vanishing of leaf regret rather than predictive loss. These findings suggest that model selection in probabilistic classification should not prioritize solely low bias, but must also account for the observational multiplicity induced by small partitions. The crossing point or elbow in the regret curve serves as a practical heuristic for identifying models that balance predictive performance with individual-level stability.

\subsection{Comparative Regret Analysis}

Using the validated framework, we analyzed the regret distribution across the three-dataset benchmark used in the decomposition validation step. Table \ref{tab:regret_results} reports the mean values for both regret components.

\begin{table}[ht]
\centering
\caption{Mean Leaf and Structural Regret across benchmark datasets.}
\label{tab:regret_results}
\begin{tabular}{p{3cm}p{2.5cm}p{3.5cm}p{1.5cm}}
\toprule
\textbf{Dataset} & \textbf{Leaf Regret} & \textbf{Structural Regret} & \textbf{Ratio} \\ \midrule
\texttt{taiwan\_credit} & 0.000096 & 0.001473 & 15.34 \\
\texttt{hmeq} & 0.001511 & 0.010521 & 6.96 \\
\texttt{bank\_marketing} & 0.000120 & 0.001520 & 12.67 \\ \bottomrule
\end{tabular}
\end{table}

Our findings, summarized in Table \ref{tab:regret_results}, reveal that \textit{structural regret} consistently dominates the total predictive uncertainty. In datasets such as \texttt{taiwan\_credit}, the structural component is more than 15 times larger than the leaf regret. This indicates that the primary driver of observational multiplicity in decision trees is the instability of the partition boundaries rather than the finite-sample noise within the leaves.

\subsection{Selective Prediction and Safety Promotion}

We evaluate how our regret decomposition can be utilized as an abstention mechanism to promote predictive safety. In high-stakes financial domains, the objective of selective prediction is to identify and abstain from ``arbitrary'' predictions where the model's output is highly sensitive to the specific realization of training labels. For this analysis, we use an expanded six-dataset benchmark that adds \texttt{german\_credit}, \texttt{loan\_data}, and \texttt{poland\_credit} to the three datasets used in the decomposition validation.

\begin{figure}[ht]
    \centering
    \includegraphics[width=\textwidth]{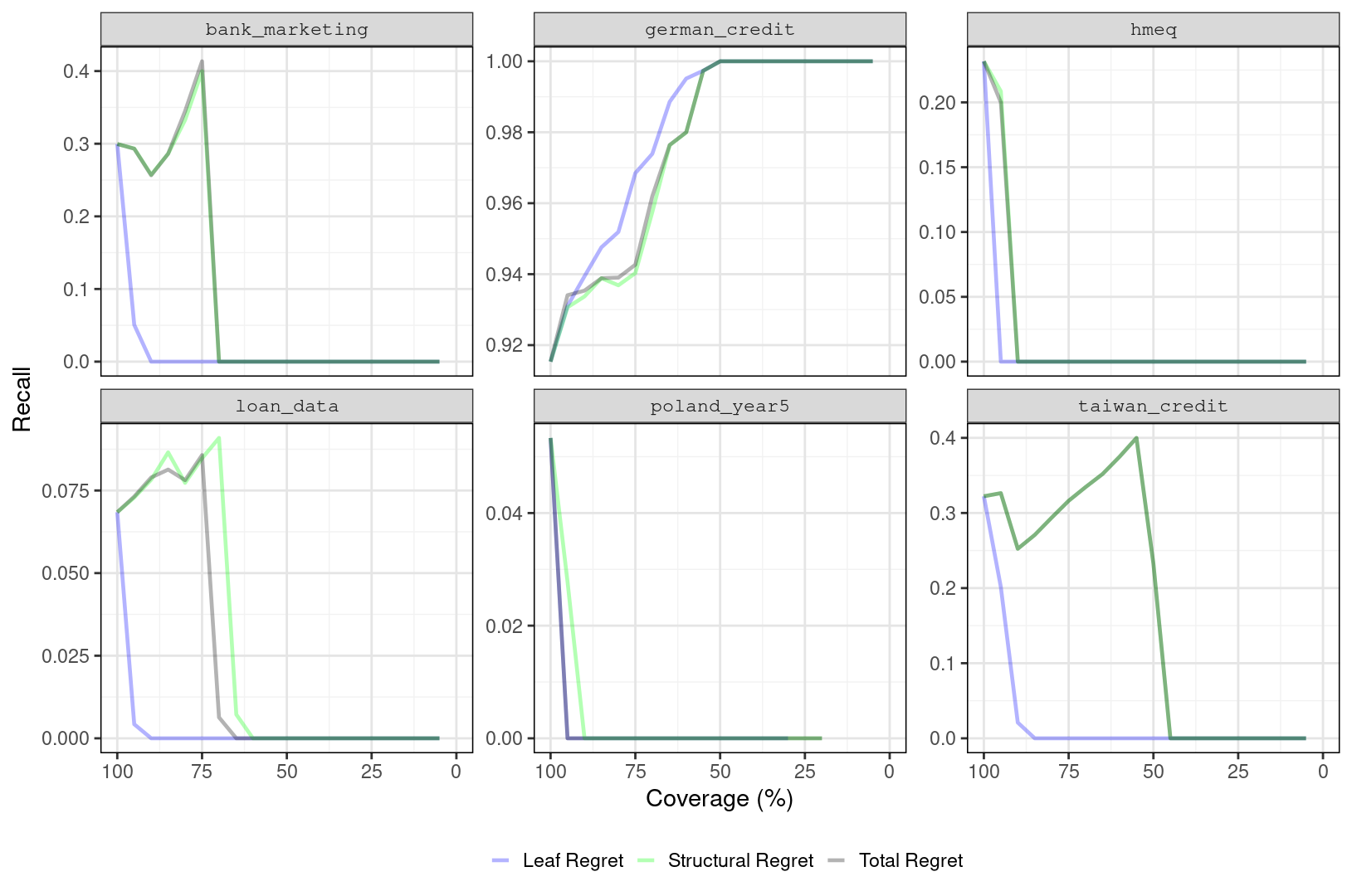}
    \caption{Selective Prediction: Recall vs. Coverage across six datasets. The x-axis is reversed, moving from full dataset utilization as 100\% to selective prediction on the most stable individuals. Curves represent ranking strategies based on Leaf Regret, Structural Regret, and Total Regret.}
    \label{fig:selective_recall}
\end{figure}

Figure~\ref{fig:selective_recall} illustrates the evolution of \textit{Recall} as a function of \textit{Coverage}, where individuals are ranked and sequentially removed based on their estimated regret.

The empirical results across six datasets reveal several critical insights regarding individual-level stability:

\begin{itemize}
    \item \textbf{Safety via Abstention:} In the \texttt{german\_credit} dataset, we observe a monotonic increase in Recall from 92\% to 100\% as coverage decreases. This confirms that ranking individuals by our proposed regret measures successfully identifies and prioritizes samples where the model's ``safety'' is highest, effectively mitigating the impact of observational multiplicity.
    
    \item \textbf{Identification of Arbitrary Regions:} In several datasets, Recall drops sharply toward zero at lower coverage levels. Rather than indicating metric failure, this behavior reflects crucial model ``honesty'': positive-class predictions in these regions are highly arbitrary and structurally unstable. The inability to maintain performance at low coverage suggests that minority-class knowledge is dominated by noise rather than signal.
    
    \item \textbf{Component Efficacy:} Consistent with our findings in Table~\ref{tab:regret_results}, \textit{Structural Regret} often provides a more robust filtering mechanism for identifying hard-to-miss positives compared to \textit{Leaf Regret}. In the \texttt{loan\_data} and \texttt{taiwan\_credit} panels, the structural component maintains predictive utility for a wider range of coverage, proving that the instability of partition boundaries is the primary informant for predictive risk.
\end{itemize}

These findings demonstrate that, by quantifying specific sources of arbitrariness, decision-makers can establish coverage thresholds that satisfy domain-specific safety requirements, ensuring that predictions are not merely accurate on average but also stable at the individual level.

\section{Discussion} 

The results presented in this study provide a granular perspective on individual-level predictive uncertainty in decision trees. By decomposing observational multiplicity into leaf and structural regret, we move beyond aggregate stability metrics and make the source of arbitrariness identifiable. Across the evaluated datasets, structural instability consistently dominates leaf-level noise, indicating that instability in partition boundaries is the principal mechanism behind prediction variability.

This interpretation has direct implications for safety-critical deployment. If high uncertainty is driven primarily by structural regret, then mitigation should prioritize model-structure stabilization (e.g., stronger regularization, pruning, or aggregation) rather than only increasing leaf sample sizes. The selective prediction results make this operational: abstaining on high-regret instances improves the reliability of positive-class identification and provides a transparent mechanism to flag cases where the model is effectively uncertain.

The observed recall--coverage behavior also supports a practical ``honesty'' principle for human-in-the-loop decision systems: when recall collapses in low-coverage regions, the model is signaling that confident prediction is not statistically supportable for those individuals. In high-stakes domains such as healthcare, credit scoring, and legal risk assessment, this signal can guide escalation to manual review instead of arbitrary automated decisions. A limitation of the present study is that it focuses on binary classification with single decision trees; extending the framework to multiclass settings and ensemble architectures remains an important direction for future work.

\section{Conclusion} 

In this paper, we formalized and validated a decomposition framework for observational multiplicity in decision trees. We introduced \textit{leaf regret} and \textit{structural regret} as complementary measures of local label variability and global partition instability, and provided theoretical guarantees for their estimation and decomposition.

Empirically, the proposed framework closely matches simulated predictive variance and shows that structural instability is the dominant source of arbitrariness. We further demonstrated a practical safety use case through selective prediction, where regret-based abstention helps identify predictions that are most sensitive to label-redrawing noise.

Overall, the framework offers a principled way to separate irreducible label uncertainty from model-structure instability, supporting more stable and justifiable individual-level decisions in high-stakes settings. Future work will extend these results to multiclass classification and ensemble tree methods.


\begin{credits}

\subsubsection{\discintname}
The authors have no competing interests to declare that are relevant to the content of this article.

\end{credits}
%
%
%
%

\end{document}